\documentclass[final]{cvpr}

\usepackage{times}
\usepackage{epsfig}
\usepackage{graphicx}
\usepackage{amsmath}
\usepackage{amssymb}
\usepackage{multirow}
\usepackage{color}
\usepackage{booktabs}
\usepackage{verbatim}

\usepackage{enumerate}
\newcommand\blfootnote[1]{%
\begingroup
\renewcommand\thefootnote{}\footnote{#1}%
\addtocounter{footnote}{-1}%
\endgroup
}

\usepackage[pagebackref=true,breaklinks=true,colorlinks,bookmarks=false]{hyperref}

\def\cvprPaperID{3805} 
\def\confYear{CVPR 2021}

\begin{document}

\title{Dynamic Scale Training for Object Detection}

\author{Yukang Chen$^{1*}$, Peizhen Zhang$^{2*}$, Zeming Li$^{2}$, \\
Yanwei Li$^{1}$, Xiangyu Zhang$^{2}$, Lu Qi$^{1}$, Jian Sun$^{2}$, Jiaya Jia$^{1}$\\
$^1$ \emph{The Chinese University of Hong Kong} $^2$ \emph{MEGVII Technology}
}

\maketitle

\begin{abstract}
    We propose a Dynamic Scale Training paradigm (abbreviated as DST) to mitigate \textit{scale variation} challenge in object detection. Previous strategies like image pyramid, multi-scale training, and their variants are aiming at preparing scale-invariant data for model optimization. However, the preparation procedure is unaware of the following optimization process that restricts their capability in handling the scale variation. Instead, in our paradigm, we use feedback information from the optimization process to dynamically guide the data preparation. The proposed method is surprisingly simple yet obtains significant gains (\textbf{2}\%+ Average Precision on MS COCO dataset), outperforming previous methods. Experimental results demonstrate the efficacy of our proposed DST method towards scale variation handling. It could also generalize to various backbones, benchmarks, and other challenging downstream tasks like instance segmentation. It does not introduce inference overhead and could serve as a free lunch for general detection configurations. Besides, it also facilitates efficient training due to fast convergence. Code and models are available at \href{https://github.com/yukang2017/Stitcher}{github.com/yukang2017/Stitcher}.
    \blfootnote{$^*$ Equal contribution. Work done during an internship in MEGVII.}
\end{abstract}



\section{Introduction}
\label{sec:intro}

 \textit{Scale variation}, a phenomenon that detection quality varies dramatically from one scale to another, originating from the imbalanced distribution of objects across different scales. It remains an unsolved challenge in object detection. In nature photography, it is impossible for an image to guarantee a balanced distribution of object patterns over different scales. Training model without handling this issue will not only depress the capability of detecting objects with minority scales but also hinder the overall performance.

\begin{figure}[htbp]
\centering
\includegraphics[width=\linewidth]{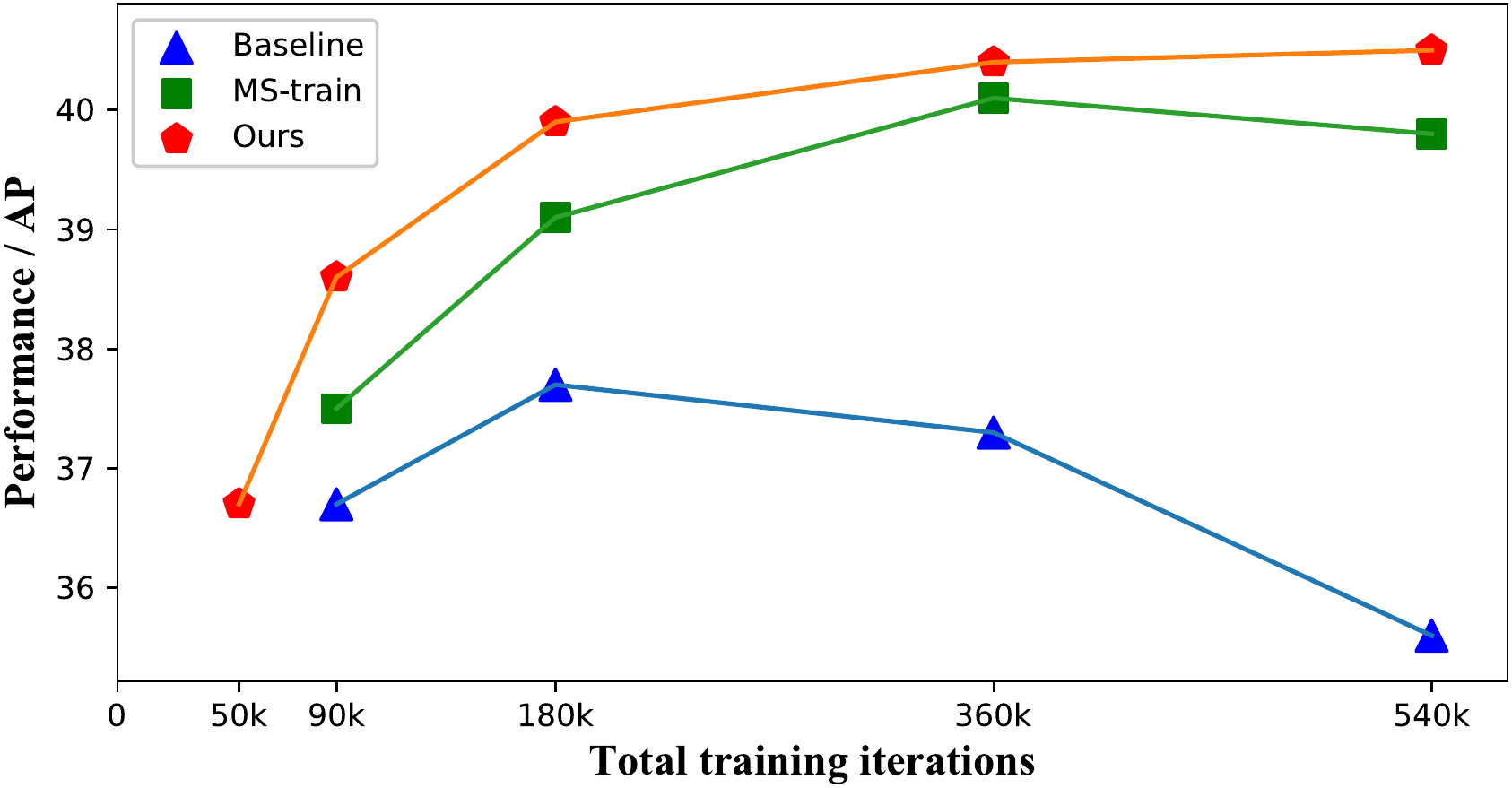}
\caption{Performance varies from baseline and multi-scale training to ours as training proceeds. Experiments are conduct on Faster R-CNN~\cite{faster-rcnn} with ResNet-50~\cite{he2016deep} FPN~\cite{fpn}. Our method consistently boosts the performance even for much longer training periods. However, the baseline and the multi-scale variant encounter severe over-fitting. Please refer to Table~\ref{tab:longerperiod} for more details.}
\label{fig:total_iters_performance}
\end{figure}

Generally, existing methods alleviate the scale variance in virtue of \textit{data preparation} or \textit{model optimization}. For instance, in the data preparation literature, image pyramid~\cite{imagepyramid} and multi-scale training augment inputs with multiple resolutions. In model optimization, feature pyramids~\cite{fpn,panet,nas-fpn} enhance representations at different receptive levels. TridentNet and POD~\cite{tridentnet,peng2019pod} propose scale-invariant architectures for assembling dilated information. Ming~\textit{et al.}~\cite{ming2019group} design architectures by balanced loss penalization. However, the above methods omit the collaboration of data preparation and model optimization. On one aspect, the preparation strategies do not fully exploit the information from model optimization, and likely to produce static augmented data blind to the dynamic optimization requirements. Also shown in Figure~\ref{fig:total_iters_performance}, strategies like multi-scale training might encounter over-fitting if persistently learning the static data patterns\footnote{Unlike \cite{he2019rethinking} that using SyncBN~\cite{megdet} and GN~\cite{wu2018group}, we fix BN~\cite{ioffe2015batch} across all experiments for common settings.}. On another aspect, the model optimization tends to be sub-optimal if no training data with desired scale information is prepared. AutoAugment~\cite{autoaug,autoaugfordet} considers  the collaboration during searching but their preparation strategy is static in the re-retraining stage. Besides, existing dynamic training methods focus on the collaboration to the label assignment, sample mining, or feature aggregation, without considering the data preparation.



In this paper, we propose a simple yet effective \textit{Dynamic Scale Training (DST)} paradigm to mitigate the scale variation issue. This is accomplished by designing a feedback-driven, dynamic data preparation paradigm to meet the optimization requirement. To resolve the requirement, we opt for tracking the penalization intensities, instantiated by loss proportions over different scales. For convenience, we adopt the loss proportion owing to the minority scale of objects as feedback. Since this statistics reflect the scale variation information of the most underwhelming samples under the background of imbalanced optimization. We deem small scale to be the minority as is acknowledged. In general, the issues to concern are (1) how to devise a handy enough data preparation strategy with potential capability towards scale variation handling  (2) how to dynamically guide this strategy, given loss proportion of small objects as feedback. For the first issue, we introduce a collage fashion of down-scaled images\footnote{other than direct re-scaling in multi-scale training that might cause extra overheads by potential large resolution of augmented data.} (see Figure~\ref{fig:image_collage}). This augmentation will potentially introduce objects with smaller sizes that might help rectify the optimization bias against majority scales (medium and large objects). Critically for the second issue, we devise a feedback-driven decision paradigm to dynamically determine the exploitation of the collage data, according to the loss statistics of the minority scales.



We experiment with our proposed DST method in various settings (backbones, training periods, and datasets). Results demonstrate that our method enhances performance consistently by handling scale variation. We also observe its versatility to different tasks by improving the performance on instance segmentation.

In summary, our contributions are two-fold:
\begin{itemize}
    \item We propose a feedback-driven, dynamic data preparation paradigm for handling scale variation.
    \item We introduce a handy collage fashion of data augmentation, which would then be guided by the feedback at runtime.
\end{itemize}


\section{Related Works}
\label{sec:relatedworks}
In this section, we shall give a brief retrospect to previous works about scale variation handling and start investigating the literature of dynamic training in object detection.

\subsection{Scale Variation Handling}
Current works for handling scale variation can be categorized into data preparation and model optimization.

\paragraph{Handling by Data Preparation}
Resampling is an intuitive method to handle scale variation, which is equivalent to amplify the loss magnitude of certain scales. However, the improvement could be limited and might hurt the performance of the other scales (see Table~\ref{tab:multismallloss}).
Image pyramid~\cite{imagepyramid} has been popular since the era of hand-crafted descriptor learning to remedy scale variation. In recent years, multi-scale training becomes common for object detection. Features learned in this way are more robust to scale variation.
However, both of the above strategies require additional overhead and storage consumption owing to transformed data with large resolutions. Moreover, since the target resolution is randomly chosen, an undesired data scale might be sub-optimal for handling scale variation.

SNIP and SNIPER~\cite{SNIP,SNIPER} are advanced versions of image pyramids. SNIP~\cite{SNIP} is proposed to normalize the object scales under multi-scale training. SNIPER~\cite{SNIPER} sample patches, instead of regular inputs for training. It meticulously crops chips around the foregrounds and backgrounds to obtain training samples. 
However, the above methods rely on multi-scale testing that suffers from inference burden. Also, their strategies are fixed as training proceeds, overlooking the dynamic merits.

Unlike above specialized methods, customized augmentations like AutoAugment~\cite{autoaug,autoaugfordet} plausibly relieve the variation problem to some extent. These methods involve thousands of GPU days for optimizing the policy controller before actual re-training. Moreover, the searched policy is also fixed during re-training without adapting the optimization.

YOLOv4~\cite{yolov4}, and Zhou et al.~\cite{cheap-pretrain} involve similar image processing to our collage fashion. We claim the novelty about this since they are concurrent works to ours. YOLOv4 use Mosaic as data augmentation. Zhou, \textit{et al.} crops foreground patches to construct jigsaw assembly for upstream classification. Instead, our method focuses on utilizing the collage images guided by dynamic feedback for handling scale variation.

\paragraph{Handling by Model Optimization}
Another line of effort for handling scale variation mainly exists in scale-invariant model optimization. This usually falls into two categories: the feature pyramids or the dilation based methods.

Feature pyramid methods aggregate information from multi-resolution levels. For instance, SSD~\cite{ssd} detects objects, taking as input the feature maps from different scales. Further, FPN~\cite{fpn} and its variants, e.g., PANet and NAS-FPN~\cite{panet,nas-fpn} fully explore path aggregation to obtain high-level semantics across all scales. However, the aggregation manner is fixed during the model learning, without considering the adjustment for better training.

On the other hand, dilation based methods adaptively enlarge the receptive fields for scale robustness. Deformable convolution networks~(DCN)~\cite{deformable} generalizes dilated convolution with flexible receptive regions. TridentNet~\cite{tridentnet} and POD~\cite{peng2019pod} combine multiple branches with various dilation rates to extract scale-sensitive representations. However, dilation based methods are not storage-friendly due to the high-resolution intermediate feature maps.

\subsection{Dynamic Training for Object Detection}
Currently, dynamic training utilized in object detection typically exists in online sample mining, feature aggregation, and label assignment. For sampling mining, OHEM~\cite{ohem} exploits region of interests (\textit{RoIs}) for hard example mining according to the cost penalization. LapNet~\cite{lapnet} introduces dynamic loss weight to indirectly conduct sample mining. For feature aggregation, FSAF~\cite{FSAF} adaptively selects the most suitable features guided by the detection loss. ASFF~\cite{ASFF} automatically learns the aggregation manner by dynamic masking. For label assignment, Liu~\textit{et al.} propose HAMBox~\cite{hambox} with dynamic compensation towards mismatched ground-truths. FreeAnchor~\cite{freeanchor} seeks for adaptive anchor-target matching during optimization. In MAL~\cite{MAL}, the number of anchors shrinks progressively as the training proceeds. ATSS~\cite{atss} proposes target-dependent training sample selection. Zhang~\textit{et al.} propose Dynamic R-CNN~\cite{dynamic-rcnn} for two-stage detectors. It progressively increases the Intersection-over-Union (\textit{IoU}) threshold for better label assignment. However, none of the above methods refer to the data preparation which is also critical to the model training. In this paper, we propose an effective feedback-driven data preparation paradigm for scale variation handling.

\section{Methodology}
\label{sec:Approach}
\begin{figure}[t]
    \centering
    \includegraphics[width=\linewidth]{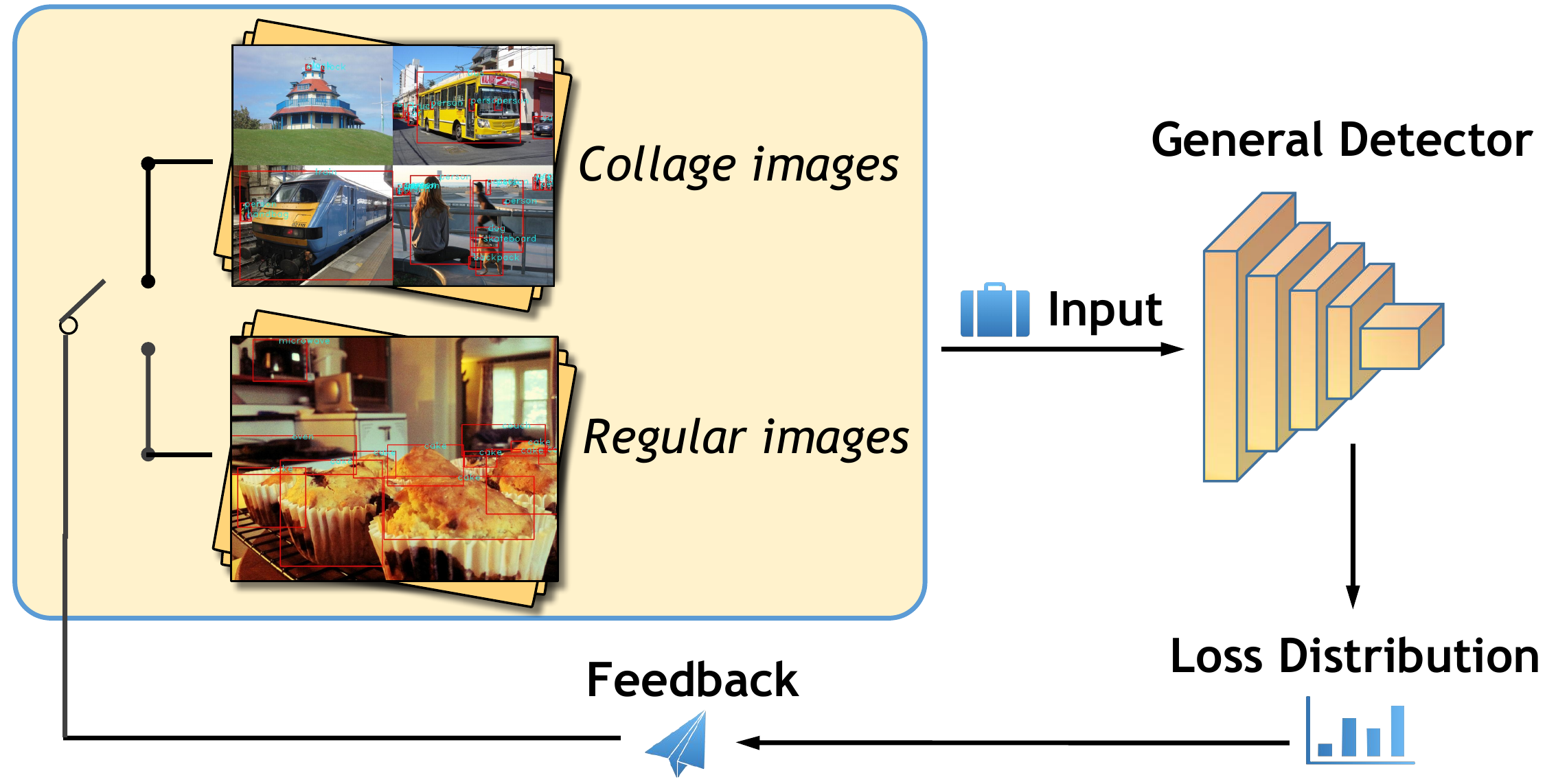}
    \caption{The pipeline of Dynamic Scale Training.}
    \label{fig:pipeline}
\end{figure}

In this section, we shall briefly give a discussion about the scale variation issue. Subsequently, we will introduce the feedback-driven data preparation paradigm followed by the collage fashion of data augmentation. The overall pipeline of the proposed dynamic scale training framework is shown in Figure~\ref{fig:pipeline}.

\subsection{A Brief Discussion about Scale Variation}
\label{sec:analysis}

\textit{Scale variation} refers to the phenomenon where models perform unfairly over different scales, featuring bad detection quality towards objects with minority scales. This commonly results from imbalanced frequencies of occurrence for instances belonging to different scales in the input images. Such imbalanced distribution would probably lead to biased network optimization. In many cases, the minority scales indicate the small scales. 

Without loss of generality, we conduct statistics upon MS COCO~\cite{coco} dataset and find two observations below:

\begin{enumerate}[(a)]
    \item \textit{Imbalance across Dataset Does Not Affect}:
    Small\footnote{we follow the scale protocol in MS COCO~\cite{coco} referred in Sec.~\ref{sec:implementation_details}. For fair annotation usage, we use the box area instead of the mask area as the size metric.} objects hold above 41\% instances in the dataset, breaking their rare stereotype. However, they still suffer from low-quality detection.
    \item \textit{Imbalance over Images that Matters}:
    medium and large objects exist in 71\% and 83\% of the images, respectively. In contrast, only around 52\% of the images contain small objects.
\end{enumerate}

Based on these observations, we believe that it is the imbalance over image distribution that leads to biased optimization towards different scales. This implies an overlooked concentration of the minority scales.

\subsection{Our Approach}
\label{sec:method}

\subsubsection{Feedback-Driven Data Preparation Paradigm}
\label{sec:method_train_level}

We propose a feedback-driven data preparation paradigm. In each training iteration, we fetch the loss proportion owing to small objects as feedback. It could be calculated after each forward propagation during model training.

Subsequently, if the loss proportion statistics is below a certain threshold in current iteration $t$, we deem it the timing to relieve imbalanced network optimization by latent compensation. In detail, we will construct collage images as input data instead of employing regular images in the next iteration $t+1$. Otherwise, if this statistic above the threshold, the regular images will serve as the input data in the coming iteration just like the default data preparation setting.
The above binary deterministic paradigm could be summarized in Eq.~(\ref{equation:2}) where $\mathrm{I}^{t+1}$ denotes the mini-batch data fed into the network at iteration $t+1$, $\mathrm{I}$ and $\mathrm{I}^c$ \textit{w.r.t.} represent the regular and collage images  in the coming iterations if applied. $r_s^t$ denotes the loss proportion accounting for small-scale objects in iteration $t$. $\tau$ is the decision threshold to control data preparation. 

\begin{equation}
\label{equation:2}
\mathrm{I}^{t+1}=\left\{
\begin{aligned}
    &\mathrm{I}^c, &if\;r_s^t \leq \tau,\\
    &\mathrm{I}\;, &otherwise.
\end{aligned}\right.
\end{equation}

From another perspective, the proposed feedback-driven paradigm could be viewed as an agent optimization, ease of policy gradients in reinforcement learning. Specifically, in the environment of object detector training, given the loss proportion observation in each training iteration, a non-parametric controller utilizes the aforementioned deterministic policy (specified in Eq.~(\ref{equation:2})) to sample from a binary action space, composed of regular or collage fashion of image processing for the next iteration of data preparation.

\subsubsection{Collage Fashion of Data Augmentation}
\label{sec:method_image_level}

\begin{figure}[htbp]
    \centering
    \includegraphics[width=1.0\linewidth]{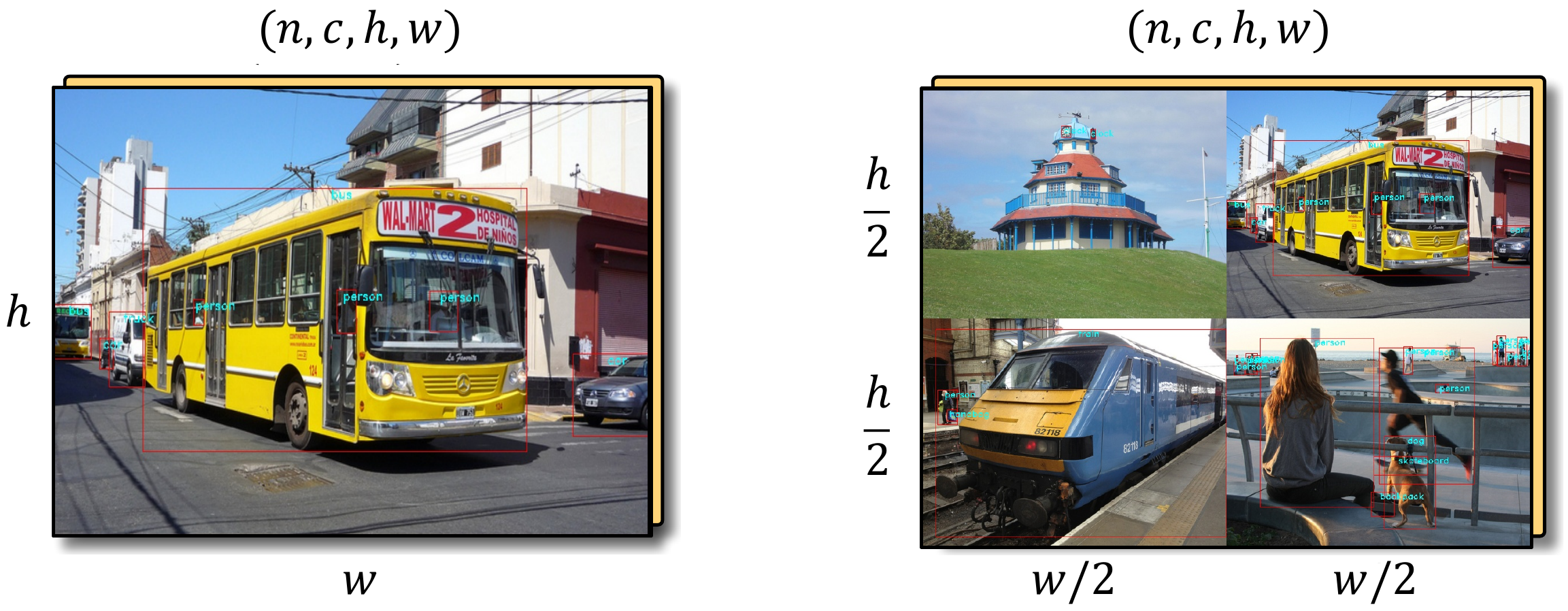}
    \caption{Regular images and collage images.}
    \label{fig:image_collage}
\end{figure}


As stated in Sec.~\ref{sec:intro}, we propose a collage fashion of scalable data augmentation for the purpose of convenient manipulation for dynamic training.

For simplicity and keeping the aspect ratio for retaining object shape priors, we formulate the collage by down-scaling and stitching $k$ regular images arranged in an equal number of rows and columns. Hence, $k$, equals to the square of row/column number, \textit{e.g.}, $1,2^2,3^2$, and so on. The spatial resolution of each component image inside is $(\frac{h}{\sqrt{k}}, \frac{w}{\sqrt{k}})$. Aside from the data, the box annotations of each source component image would get properly rescaled and translated for consistency. When $k$ equal to 1, a collage image degenerates to a regular image.

Figure~\ref{fig:image_collage} shows a collage (right) specifying $k=4$ compared to a regular image (left). 

 As can be seen, the collage fashion of image processing introduces a minimal scale variation handling by explicitly manufacturing object patterns with smaller scales. And Since collage images retain identical size as regular images, no additional overhead involves in network propagation.

\section{Experiments}

\label{sec:experiments}

In this section, we begin by briefly describing the implementation details. Whereafter, the efficacy analysis of our proposed method compared to previous works are investigated. Next, we shall elaborate on the ablation studies. A quantitative analysis of the scale variation issue will also be given. We end the experiment section by discussing extra merits and corner cases brought by the proposed method.

\subsection{Implementation Details}
\label{sec:implementation_details}

Experiments are mainly conducted on the challenging MS COCO~\cite{coco} dataset which contains 80 categories. Following the common practice in \cite{R-CNN}, the union of the primitive training set (80k images) and the {\fontfamily{qcr}\selectfont trainval35k} subset (35k images) of primitive validation set are used for training. The evaluation is conducted on the {\fontfamily{qcr}\selectfont minival} subset with 5k images. We follow the scale protocol in COCO to distinguish the small, middle and large objects by 32$^2$ and 96$^2$ pixel areas. Input images are resized such that their shorter side is 800 and the longer side no more than 1,333. 

Throughout all experiments, the initial learning rate is set as 0.02 with Stochastic Gradient Descent (SGD) with momentum as 0.9 and weight decay as 1e-4. The mini-batch size is set to 16 (2 images per GPU). The network is trained for 90k iterations that will be decayed by 10 at 60k and 80k iterations, respectively. For longer training periods if required, we adopt a common proportional milestones extension. For example, a 2$\times$ setting with 180k iterations, and milestones at 120k and 160k respectively.

Besides MS COCO, we also examine our efficacy of handling scale variation on PASCAL VOC~\cite{PascalVOC} dataset. Moreover, extra studies on the challenging instance segmentation task also verify the versatility of our proposed method.

\subsection{Comparison to Previous Methods}

\subsubsection{Comparison to Resampling}
\label{sec:compare_resampling}
Following the spirit of the resampling strategy for more balanced training, we apply a careful re-weight scheme to assist the minority scales in each iteration. In detail, we amplify the loss magnitudes of small objects to be equal to that of the medium and large objects. However, as shown in Table~\ref{tab:multismallloss}, the overall performance and the performance of the other scales deteriorate with an only slight improvement to the small scale (AP$_s$ +0.3\%).

\begin{table}[htbp]
    \centering
    \caption{Impact brought by resampling.}
    \label{tab:multismallloss}
    \resizebox{0.75\linewidth}{!}{
    \begin{tabular}{l|cccc}
    \toprule
                        & AP & AP$_s$ & AP$_m$  & AP$_l$ \\
    \midrule
    Baseline  & 36.7 & 21.1 & 39.9 & 48.1 \\
    + Resampling & 36.4 & 21.4 & 39.3 & 47.4 \\
    \bottomrule
    \end{tabular}}
\end{table}


\begin{table*}[h]
\caption{Comparison with common baselines and multi-scale training on Faster R-CNN.}
\label{tab:faster-rcnn-1x}
\centering
\resizebox{0.7\linewidth}{!}{
\begin{tabular}{l|l|c|ll|ll|ll|ll}
\toprule
              & \multicolumn{1}{c|}{Backbone}                        & Hours & \multicolumn{2}{c|}{AP} & \multicolumn{2}{c|}{AP$_s$} & \multicolumn{2}{c|}{AP$_m$} & \multicolumn{2}{c}{AP$_l$} \\ \midrule
Baseline      & \multirow{5}{*}{ResNet-50 FPN}  & 8.7   & 36.7           &        & 21.1             &          & 39.9             &          & 48.1             &          \\
MS-train$^{s}$    &                                 &  8.1     & 36.3           &        & 23.7             & \textcolor{blue}{(+2.6)}   & 39.9             &          & 45.9             & \textcolor[rgb]{0,0.4,0.2}{(-2.2)}   \\
MS-train$^{m}$    &                                 & 10.8  & \textbf{37.5}           &        & 22.0             &          & 40.7             &          & 48.8             &          \\
MS-train$^{l}$    &                                 & 14.4  & 37.1           &        & 20.7             & \textcolor[rgb]{0,0.4,0.2}{(-0.4)}   & 40.3             &          & 49.8    & \textcolor{blue}{(+1.7)}   \\
\textbf{Ours} &                                 & 9.0   & \textbf{38.6}  & \textbf{(+1.9)} & \textbf{24.4}    & \textbf{(+3.3)}   & 41.9    & (+2.0)   & 49.3             & (+1.2)   \\ \midrule
Baseline      & \multirow{5}{*}{ResNet-101 FPN} & 11.5  & 39.1           &        & 22.6             &          & 42.9             &          & 51.4             &          \\
MS-train$^{s}$    &                                 & 10.8  & 38.9           &        & 24.2             & \textcolor{blue}{(+1.6)}   & 42.7             &          & 49.0             & \textcolor[rgb]{0,0.4,0.2}{(-2.4)}   \\
MS-train$^{m}$    &                                 & 14.2  & \textbf{39.7}           &        & 23.6             &          & 43.3             &          & 51.3             &          \\
MS-train$^{l}$    &                                 & 21.3     & 39.3           &        & 22.3             & \textcolor[rgb]{0,0.4,0.2}{(-0.3)}   & 43.0             &          & 51.9    & \textcolor{blue}{(+0.5)}   \\
\textbf{Ours} &                                 & 11.7  & \textbf{40.8}  & \textbf{(+1.7)} & \textbf{25.8}    & \textbf{(+3.2)}   & 44.1    & (+1.2)   & 51.9    & (+0.5)   \\ \bottomrule
\end{tabular}}
\end{table*}

\begin{table*}[h]
\caption{Comparison with common baselines and multi-scale training on Faster R-CNN for 2x training periods.}
\label{tab:faster-rcnn-2x}
\centering
\resizebox{0.72\linewidth}{!}{
\begin{tabular}{l|l|c|ll|ll|ll|ll}
\toprule
              & \multicolumn{1}{c|}{Backbone}                        & Hours & \multicolumn{2}{c|}{AP}         & \multicolumn{2}{c|}{AP$_s$}     & \multicolumn{2}{c|}{AP$_m$} & \multicolumn{2}{c}{AP$_l$} \\ \midrule
Baseline      & \multirow{3}{*}{ResNet-50 FPN}  & 17.2  & 37.7          &                 & 21.6          &                 & 40.6        &               & 49.6        &               \\
MS-train$^{m}$    &                                 & 20.5  & 39.1          &                 & 23.5          &                 & 42.2        &               & 50.8        &               \\
\textbf{Ours} &                                 & 17.5  & \textbf{39.9} & \textbf{(+2.2)} & \textbf{25.1} & \textbf{(+3.5)} & 43.1        & (+2.5)        & 51.0        & (+1.4)        \\ \midrule
Baseline      & \multirow{3}{*}{ResNet-101 FPN} & 23.4  & 39.8          &                 & 22.9          &                 & 43.3        &               & 52.6        &               \\
MS-train$^{m}$    &                                 & 28.5  & 41.6          &                 & 25.5          &                 & 45.3        &               & 54.1        &               \\
\textbf{Ours} &                                 & 23.5  & \textbf{42.1} & \textbf{(+2.3)} & \textbf{26.9} & \textbf{(+4.0)} & 45.5        & (+2.2)        & 54.1        & (+1.5)        \\ \bottomrule
\end{tabular}}
\end{table*}

\subsubsection{Comparison to Common Baselines}
\label{sec:compare_baselines}

As shown in Table~\ref{tab:faster-rcnn-1x}, the improvement against baseline is highlighted in parenthesis. We observe decent improvement overall ($1.7\%$+ AP), and more significant results for the minority scales, \textit{i.e.}, the small scales ($\mathbf{3.2}$\%+ AP$_s$). Table~\ref{tab:faster-rcnn-2x} shows the comparison in 2$\times$ training periods, presenting even higher gains ($2.2$\%+ AP and up to $\mathbf{4.0}$\% AP$_s$).


We also conduct counterpart experiments on single stage detectors, \textit{e.g.}, RetinaNet~\cite{retinanet} and FCOS~\cite{tian2019fcos} as shown in Table~\ref{tab:retinanet_fcos}. 

These demonstrate the effectiveness not only on general detection enhancing but also on scale variation handling \textit{esp.} for the minority scales using dynamic scale training.


\subsubsection{Comparison to Multi-scale Training}
\label{sec:compare_multiscale}

\paragraph{(a) Different settings of multi-scale training} \textbf{} \\
We carefully compare our method with multi-scale training with various scale settings as exhibited in Table~\ref{tab:faster-rcnn-1x}. Here, MS-train$^{s}$, MS-train$^{m}$ and MS-train$^{l}$ correspond to sampling intervals about the shorter side length, denoted as \underline{[400, 800]}, \underline{[600, 1000]}, and \underline{[800, 1200]} respectively, with stride 100. They indicate settings prefer to small, middle, and large scale respectively. Among them, Multi-scale$^m$ achieves the best trade-off, as the other two settings acquire improvement in their favorite scale at the price of greatly harming the opposite scale (highlighted in blue and green in Table~\ref{tab:faster-rcnn-1x}). Hence, We adopt the Multi-scale$^m$ setting for Multi-scale training in the following experiments. Yet, our method still outperforms this strategy across all scales.


\paragraph{(b) Time efficiency} \textbf{} \\
The proposed dynamic scale training method brings about negligible overhead compared to baselines. It mainly comes from the collage augmentation which involves $nearest$ neighbor interpolation for down-scaling component images. Empirically, a collage operation costs about 0.02 seconds in a single training iteration. Since the frequencies of collage operation depend on the dynamic preparation paradigm that is unavailable in advance. We practically measure the time consumption in terms of the complete training period. All measurements are benchmarked on 8 RTX 2080Ti GPU cards with 16 mini-batch size.

As shown in Table~\ref{tab:faster-rcnn-1x}, it takes 8.7 hours to train the baseline with ResNet-50 FPN in 1$\times$ period. Instead, multi-scale training requires extra 2 hours (10.8). The gap enlarges when experimenting on a larger backbone (ResNet-101 FPN) or longer training period (2$\times$). In contrast, our method takes only a bit longer than the baseline (9 hours with extra 0.3 hours). And the gap is invariant to the training periods (nearly the same in both 1$\times$ and 2$\times$ settings). Moreover, the gap shrinks when taking larger backbones (ResNet-101 FPN) for experiments. Please refer to Table~\ref{tab:faster-rcnn-1x} and Table~\ref{tab:faster-rcnn-2x} for details. Therefore, our proposed method is much more efficient than multi-scale training.

\begin{table}[htbp]
    \caption{Evaluation on the effect of multi-scale testing.}
    \label{tab:compare_multiscaletesting}
    \centering
    \resizebox{0.7\linewidth}{!}{
    \begin{tabular}{l|ll|l|l|l}
    \toprule
    \multicolumn{1}{l|}{} & \multicolumn{2}{c|}{AP} & \multicolumn{1}{c|}{AP$_s$} & \multicolumn{1}{c|}{AP$_m$} & \multicolumn{1}{c}{AP$_l$} \\ \midrule
    MS-train$^{m}$    & 37.5           &        & 22.0                        & 40.7                        & 48.8                       \\
    + MS-test$^{m}$   & 38.8           & (+1.3) & 23.7                        & 41.6                        & 49.8                       \\ \midrule
    Ours                  & 38.6           &        & 24.4                        & 41.9                        & 49.3                       \\
    + MS-test$^{m}$   & \textbf{39.9}  & (+1.3) & 26.5                        & 42.7                        & 51.0                       \\ \bottomrule
    \end{tabular}
    }
\end{table}

\paragraph{(c) Compatible to multi-scale testing} \textbf{}\\
It is acknowledged that models trained with multi-scale training could further enhance the performance with matching multi-scale testing. Thus, without loss of generality, we conduct a comparison by applying MS-test$^{m}$ to MS-train$^{m}$ and our proposed method, respectively. As shown in Table~\ref{tab:compare_multiscaletesting}, our proposed method shares exactly the same merit (+1.3\%). This reveals good compatibility.

\begin{table}[htbp]
    \caption{Evaluation on longer training periods.}
    \label{tab:longerperiod}
    \centering
    \resizebox{0.85\linewidth}{!}{
    \begin{tabular}{c|c|l|ccc}
    \toprule
                              & Iterations & AP            & AP$_s$  & AP$_m$  & AP$_l$  \\ \midrule
    \multirow{4}{*}{Baseline} & 90k     & 36.7          & 21.1 & 39.8 & 48.1 \\
                              & 180k     & 37.7       & 21.6 & 40.6 & 49.6 \\
                              & 360k     & 37.3 $\downarrow$         & 20.3 & 39.6 & 50.1 \\
                              & 540k     & 35.6 $\downarrow$          & 19.8 & 37.7 & 47.6 \\ \midrule
    \multirow{2}{*}{MS-train} & 90k     & 37.5 & 22.0 & 40.7 & 48.8 \\
                              & 180k     & 39.1 & 23.5 & 42.2 & 50.8 \\
                              & 360k     & 40.1 & 24.3 & 43.3 & 52.4 \\
                              & 540k     & 39.8 $\downarrow$ & 24.1 & 43.0 & 52.0 \\\midrule
    \multirow{2}{*}{Ours} & 90k     & 38.6  & 24.4 & 41.9 & 49.3 \\
    & 180k     &     39.9      & 25.1 & 43.1 & 51.0 \\
    & 360k     &     40.4      & 25.2 & 43.6 & 51.9 \\
                              & 540k     & \textbf{40.5} $\uparrow$ &  26.1 & 43.2 & 51.6 \\ \bottomrule
    \end{tabular}}
\end{table}

\begin{table*}[htbp]
    \caption{Comparison on RetinaNet and FCOS with ResNet-50 and ResNet-101 backbones for 2$\times$ training periods.}
    \label{tab:retinanet_fcos}
    \centering
    \resizebox{0.75\linewidth}{!}{
    \begin{tabular}{l|c|l|ll|ll|ll|ll}
    \toprule
                      & \multicolumn{1}{c|}{Model} & \multicolumn{1}{c|}{Backbone} & \multicolumn{2}{c|}{AP}  & \multicolumn{2}{c|}{AP$_s$}  & \multicolumn{2}{c|}{AP$_m$} & \multicolumn{2}{c}{AP$_l$} \\ \midrule

    
    Baseline          & \multirow{4}{*}{RetinaNet}         & \multirow{2}{*}{ResNet-50 FPN}  & 36.8         &            & 20.2      &               & 40.0       & & 49.7        &               \\
    Ours &            & & \textbf{39.0} & \textbf{(+2.2)} & \textbf{23.4} & \textbf{(+3.2)} & 42.9     & (+2.9)     & 51.0      & (+1.2)       \\ \cline{1-1} \cline{3-11} 
    Baseline          & & \multirow{2}{*}{ResNet-101 FPN} & 38.8          &                 & 21.1     &            & 42.1      &              & 52.4        &               \\
    Ours &            & & \textbf{41.3} & \textbf{(+2.5)} & \textbf{25.4} & \textbf{(+4.3)} & 45.1     & (+3.0)     & 54.0      & (+1.6)       \\ \midrule
    
    Baseline          & \multirow{4}{*}{FCOS}         & \multirow{2}{*}{ResNet-50 FPN}  & 37.1         &            & 21.6      &               & 41.0       & & 47.3        &               \\
    Ours &            & & \textbf{39.8} & \textbf{(+2.7)} & \textbf{25.4} & \textbf{(+3.8)} & 43.9     & (+2.9)     & 50.2      & (+2.9)       \\ \cline{1-1} \cline{3-11} 
    Baseline          & & \multirow{2}{*}{ResNet-101 FPN} & 39.1          &                 & 22.2     &            & 43.4      &              & 50.6        &               \\
    Ours &            & & \textbf{41.6} & \textbf{(+2.5)} & \textbf{26.1} & \textbf{(+3.9)} & 45.5     & (+2.1)     & 53.3      & (+2.7)       \\
    \bottomrule
\end{tabular}}
\end{table*}

\paragraph{(d) Longer training periods} \textbf{}\\
Recalling the proposed collage augmentation, one association with multi-scale training is that they both create scalable instance patterns to some extent.  However, we are wondering if multi-scale training is capable of mitigating the performance gap to ours, if long enough training periods are allowed.

To resolve this, we conduct experiments upon Faster R-CNN with ResNet-50 and FPN on various training periods as shown in Table~\ref{tab:longerperiod}. We find that the gap starts shrinking when the training process reaches sufficiently longer periods (3$\times$ to 4$\times$). However, interestingly for the longest 6$\times$ training period (540k iterations), the performance of the multi-scale training (also the baseline) encounter degradation. Instead, our method could further enhance the performance. One reasonable explanation is that the feedback-driven preparation paradigm consistently provides data of the desired scale to effectively avoid {\it over-fitting}.

\begin{table}[htbp]
    \caption{Comparison with SNIP / SNIPER.}
    \label{tab:compare_sniper}
    \centering
    \resizebox{0.9\linewidth}{!}{
    \begin{tabular}{l|l|l|ccc}
    \toprule
           & \multicolumn{1}{c|}{Backbone}                    & \multicolumn{1}{c|}{AP}        & AP$_s$ & AP$_m$ & AP$_l$ \\ \midrule
    SNIP   & \multirow{3}{*}{ResNet-50 C4}  & 43.6                           & 26.4   & 46.5   & 55.8   \\
    SNIPER &                             & 43.5                           & 26.1   & 46.3   & 56.0   \\
    Ours   &                             & \textbf{44.2} & \textbf{28.7}   & \textbf{47.2}   & \textbf{58.3}   \\ \midrule
    SNIP   & \multirow{3}{*}{ResNet-101 C4} & 44.4                           & 27.3   & 47.4   & 56.9   \\
    SNIPER &                             & 46.1                           & 29.6   & 48.9   & 58.1   \\
    Ours   &                             & \textbf{46.9} & \textbf{30.9}   & \textbf{50.5}   & \textbf{60.9}   \\ \bottomrule
    \end{tabular}}
\end{table}

\subsubsection{Comparison to SNIP and SNIPER}\label{sec:compare_sniper}
As shown in Table~\ref{tab:compare_sniper}, we compare our method to SNIP~\cite{SNIP} and SNIPER~\cite{SNIPER}{\color{red} \footnote{For fair comparisons, we use the same augmentations (deformable convolution, MS test, and soft-NMS~\cite{soft-nms}) as SNIP and SNIPER do.}} methods on various backbones. As a result, our method performs better. This might because the SNIP and SNIPER operate in a static manner during training, rendering them unable to provide scale-sensitive data that the network desires. In contrast, our method benefits from the dynamic data preparation paradigm to meet the requirements training-dependently. Moreover, our method is simpler to use while SNIPER involves extended label assignment and chips sampling procedure.

\begin{table}[htbp]
    \caption{Evaluation on Large Backbones.}
    \label{tab:largebackbone}
    \centering
    \resizebox{0.9\linewidth}{!}{
    \begin{tabular}{l|c|l|ccc}
    \toprule
                              & Backbone    & \multicolumn{1}{c|}{AP} & AP$_s$        & AP$_m$        & AP$_l$        \\ \midrule
    Baseline       & \multirow{2}{*}{ResNext 101} & 41.6                    & 24.8          & 45.1          & 53.3          \\
    Ours                                   &      & \textbf{43.1}           & \textbf{28.0} & \textbf{46.7} & \textbf{54.2} \\ \midrule
    Baseline  & \multirow{2}{*}{ResNet 101 + DCN} & 42.3                    & 24.8          & 46.1          & 55.7          \\
    Ours                                   &      & \textbf{43.3}           & \textbf{27.1} & \textbf{47.0} & \textbf{56.0} \\ \midrule
    Baseline & \multirow{2}{*}{ResNext 101 + DCN} & 44.1                    & 26.8          & 47.5          & 57.8          \\
    Ours                                   &      & \textbf{45.4}           & \textbf{29.4} & \textbf{48.8} & \textbf{58.5} \\ \bottomrule
    \end{tabular}}
\end{table}

\subsubsection{Evaluation on Large Backbones}
Table~\ref{tab:largebackbone} shows the improvement from our method on large backbones, \textit{i.e.}, ResNext 101~\cite{resnext}, ResNet-101 with DCN~\cite{deformable} and ResNext-32$\times$8d-101 with DCN~\cite{deformable}.
Based on the strong baselines, our method could still enhance the performance by 1.0\% to 1.5\% AP.

\begin{table}[htbp]
    \caption{Evaluation on Instance Segmentation.}
    \label{tab:instanceseg}
    \centering
    \resizebox{0.9\linewidth}{!}{
    \begin{tabular}{l|c|l|l|l|l}
    \toprule
             & Backbone                     & \multicolumn{1}{c|}{AP} & \multicolumn{1}{c|}{AP$_s$} & \multicolumn{1}{c|}{AP$_m$} & \multicolumn{1}{c}{AP$_l$} \\ \midrule
    Baseline & \multirow{2}{*}{ResNet-50 FPN}  & 34.3                    & 15.8                        & 36.7                        & 50.5                       \\
    Ours     &                              & \textbf{35.1}           & \textbf{17.0}               & \textbf{37.8}               & \textbf{51.4}              \\ \midrule
    Baseline & \multirow{2}{*}{ResNet-101 FPN} & 35.9                    & 15.9                        & 38.9                        & 53.2                       \\
    Ours     &                              & \textbf{37.2}           & \textbf{19.0}               & \textbf{40.3}               & \textbf{53.7}              \\ \bottomrule
    \end{tabular}}
    \end{table}

\subsubsection{Evaluation on Instance Segmentation}
Beyond object detection, we also apply our method to instance segmentation task.
Experiments are conducted on the COCO instance segmentation track~\cite{coco}. We report COCO mask AP on the {\fontfamily{qcr}\selectfont minival} split. Models are trained for 90k iterations and divided by 10 at {60k and 80k} iterations. 
We train Mask R-CNN~\cite{maskrcnn} models with Stochastic Gradient Descent (SGD), 0.9 momentum and 1e-4 weight decay and 16 batch size (2 images for per GPU).
As shown in Table~\ref{tab:instanceseg}, our method improves AP by 0.9\% on ResNet-50 and by 1.3\% on ResNet-101.

\begin{table*}[htbp]
    \caption{Evaluation on PASCAL VOC dataset on Faster R-CNN.}
    \label{tab:pascal_voc}
    \centering
    \resizebox{\linewidth}{!}{
    \begin{tabular}{l|c|cccccccccccccccccccc}
    \toprule
    & mAP  & plane & bike & bird & boat & bottle & bus  & car  & cat  & chair & cow  & table & dog  & horse & bike & person & plant & sheep & sofa & train & tv \\ \midrule
    Baseline & 80.3 & 86.9      & 86.7    & 80.1 & 72.5 & 71.9   & 86.9 & 88.4 & 88.7 & 63.3  & 87.0 & 75.3        & 88.5 & 88.4  & 80.1      & 85.5   & 56.7        & 78.2  & 78.8 & 85.0  & 77.6      \\ \midrule
    Ours & 82.6 & 89.0      & 86.7   & 80.2 & 73.0 & 72.7   & 87.0 & 89.3 & 89.0 & 68.6  & 86.8 & 79.7        & 88.8 & 88.5  & 88.1      & 87.3   & 59.8        & 86.7  & 80.2 & 88.1  & 84.0  \\
    \bottomrule
    \end{tabular}}
\end{table*}

\subsubsection{Evaluation on PASCAL VOC}
Besides MS COCO, we also generalize our proposed dynamic scale training method to Pascal VOC~\cite{PascalVOC} dataset. Following the protocol in~\cite{fast-r-cnn}, the union of {\fontfamily{qcr}\selectfont 2007 trainval} and {\fontfamily{qcr}\selectfont 2012 trainval} are used for training. Models are trained by 24k iterations in which the learning rate is set as 0.01 and 0.001 in the first two-thirds and the remaining one-third iterations, respectively. Evaluation is performed on {\fontfamily{qcr}\selectfont 2007 test}.  As shown in Table~\ref{tab:pascal_voc}, our method obtains a gain of 2.3\% mAP overall. In addition, the detection quality of small scale categories like bottle, chair, and tv get significantly improved. 

\subsection{Ablation Studies}

In this section, we analyze the best practice of the feedback choice and the deterministic threshold $\tau$ in the feedback-driven data preparation paradigm. Besides, we conduct a simple ablation on selecting the number of component images $k$ in the collage fashion. We use Faster R-CNN with ResNet-50 and FPN for the studies.

\paragraph{Feedback choice.}
To explore the preparation paradigm, we set up below control experiments as shown in Table~\ref{tab:ablation1}.

\noindent
$\centerdot$ \textit{All collage}: collage images all the time; \\
$\centerdot$ \textit{All regular}: regular images all the time (baseline); \\
$\centerdot$ \textit{Random sampling}: collage or regular images randomly; \\
$\centerdot$ \textit{Input feedback}: occurrence frequency of small instances in the input as feedback; \\
$\centerdot$ \textit{Classification/Regression/Joint loss feedback}: loss proportion of small objects as feedback.

\begin{table}[h]
    \caption{Ablation study on feedback choice.}
    \label{tab:ablation1}
    \centering
    \resizebox{\linewidth}{!}{
    \begin{tabular}{ll|c|ccc}
    \toprule
    \multicolumn{2}{c|}{feedback strategy (if any)}                                        & AP & AP$_s$ & AP$_m$ & AP$_l$ \\ \midrule
    \multicolumn{1}{l|}{\multirow{3}{*}{No}}   & All collaged            & 32.1          & 21.9  & 36.4  & 36.8  \\
    \multicolumn{1}{l|}{}                               & All regular              & 36.7          & 21.1  & 39.8  & 48.1  \\
    \multicolumn{1}{l|}{}                               & Random sampling           & 37.8          & 23.6  & 40.7  & 46.7  \\ \midrule
    \multicolumn{1}{l|}{\multirow{4}{*}{Yes}} & Input Ratio          & 38.1          & 23.1  & 41.3  & 49.1  \\
    \multicolumn{1}{l|}{}                               & Classification Loss & 38.5          & 23.9  & 41.6  & 48.8  \\
    \multicolumn{1}{l|}{}                               & Regression Loss     & 38.6 & 24.4  & 41.9  & 49.3  \\
    \multicolumn{1}{l|}{}                               & Joint Loss     & 38.5          & 23.7  & 41.6  & 49.3  \\ \bottomrule
    \end{tabular}}
\end{table}

As shown in Table~\ref{tab:ablation1}, static usage of collage images leads to bad performance. It might run into another extreme situation where learning biases towards small-scales. Besides, random sampling performs better than the common baseline, but it is still static. The dynamic feedback strategies, \textit{e.g.}, the input feedback, results in better performance. However, such input-guided feedback is inferior to the loss-guided ones since it does not consider the optimization process. Results with different loss-guided feedback strategies are comparable, robust to specific supervision tasks. By default, we use regression loss-guided for convenience.

\begin{figure}[htbp]
\centering
\includegraphics[width=0.85\linewidth]{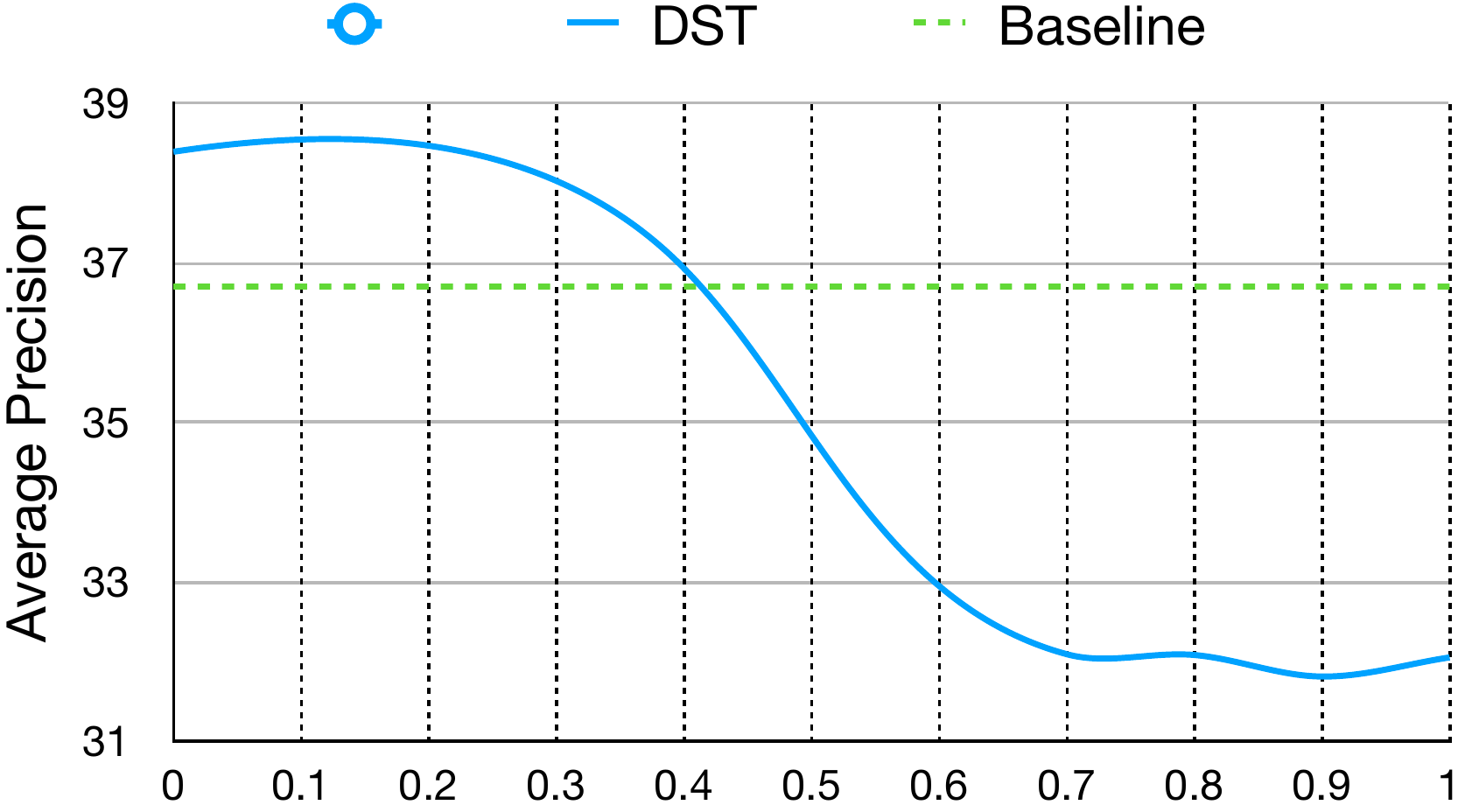}
\caption{Ablation study on the threshold $\tau$.}
\label{fig:threshold}
\end{figure}
\paragraph{Deterministic threshold.}
In the proposed method, only one hyper-parameter $\tau$ requires tuning. We apply grid searching and study the impact as shown in Figure~\ref{fig:threshold}. The performance decreases dramatically as $\tau$ exceeds 0.2. Empirically, we set $\tau$ as 0.1 and apply it across all experiments without loss of generality. Notably, it happens to be coincident with the ratio observation covering  half of the training iterations, as described in Figure~\ref{fig:loss_iterations}. This provides a promising heuristics for convenient tuning by first calculating the statistics during the baseline training on a minimal subset.

\begin{table}[htbp]
\caption{Ablation study on number of collage components.}
\label{tab:collage_component_k}
\centering
\begin{tabular}{c|c|ccccc}
\toprule
$k$     & AP            & AP$_{50}$ & AP$_{75}$ & AP$_s$ & AP$_m$ & AP$_l$ \\ \hline
$1^2$ & 36.7 & 58.4 & 39.6 & 21.1 & 39.8 & 48.1 \\ 
$2^2$ & 38.6 & 60.5 & 41.8 & 24.4 & 41.9 & 49.3 \\
$3^2$ & 38.4 & 60.5 & 41.5 & 24.2 & 41.7 & 48.8\\
\bottomrule
\end{tabular}
\end{table}

\paragraph{Number of collage components.}
We conduct a simple ablation on different number $k$ of component images used in collage by our proposed method. Since we mainly focus on the dynamic preparation paradigm, we simply adopt $k=4$ for a good trade-off as shown in Table~\ref{tab:collage_component_k}.

\subsection{Analysis of Scale Variation}
\begin{figure}[htbp]
\centering
\includegraphics[width=1.0\linewidth]{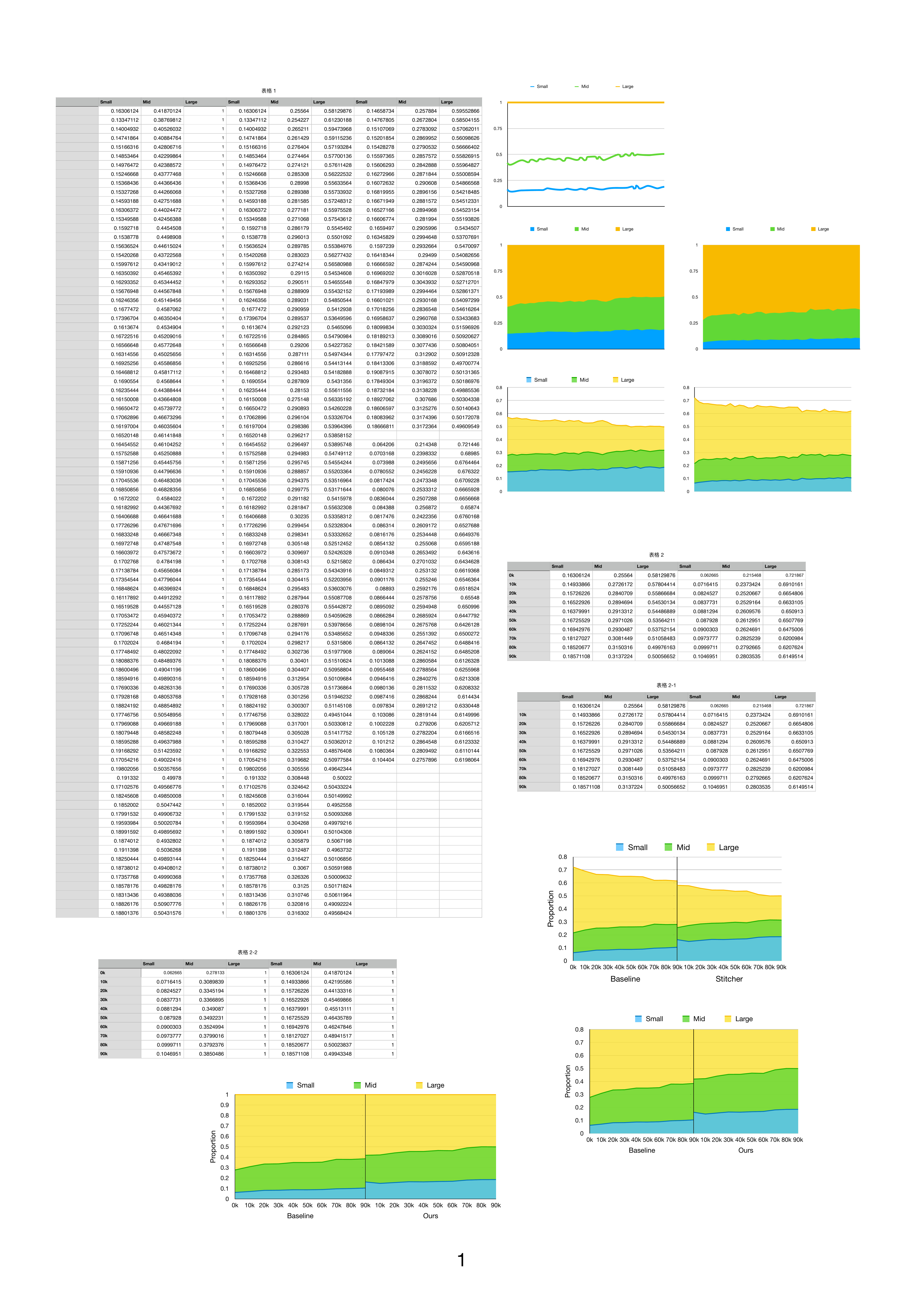}
\caption{Loss proportion across different scales before and after.}
\label{fig:loss_iterations}
\end{figure}

Besides reflecting the improvement of scale variation handling by performance gains, we also investigate in the view of optimization preference. We measure this by loss proportions occupied by different scales over iterations. These statistics are collected from the training process of Faster R-CNN with ResNet-50 and FPN. As a result, we draw the curves of model training w/ and w/o our proposed method in Figure~\ref{fig:loss_iterations}. It can be observed that the scale variation gets much alleviated.

\begin{figure}[htbp]
    \centering
    \includegraphics[width=1.0\linewidth]{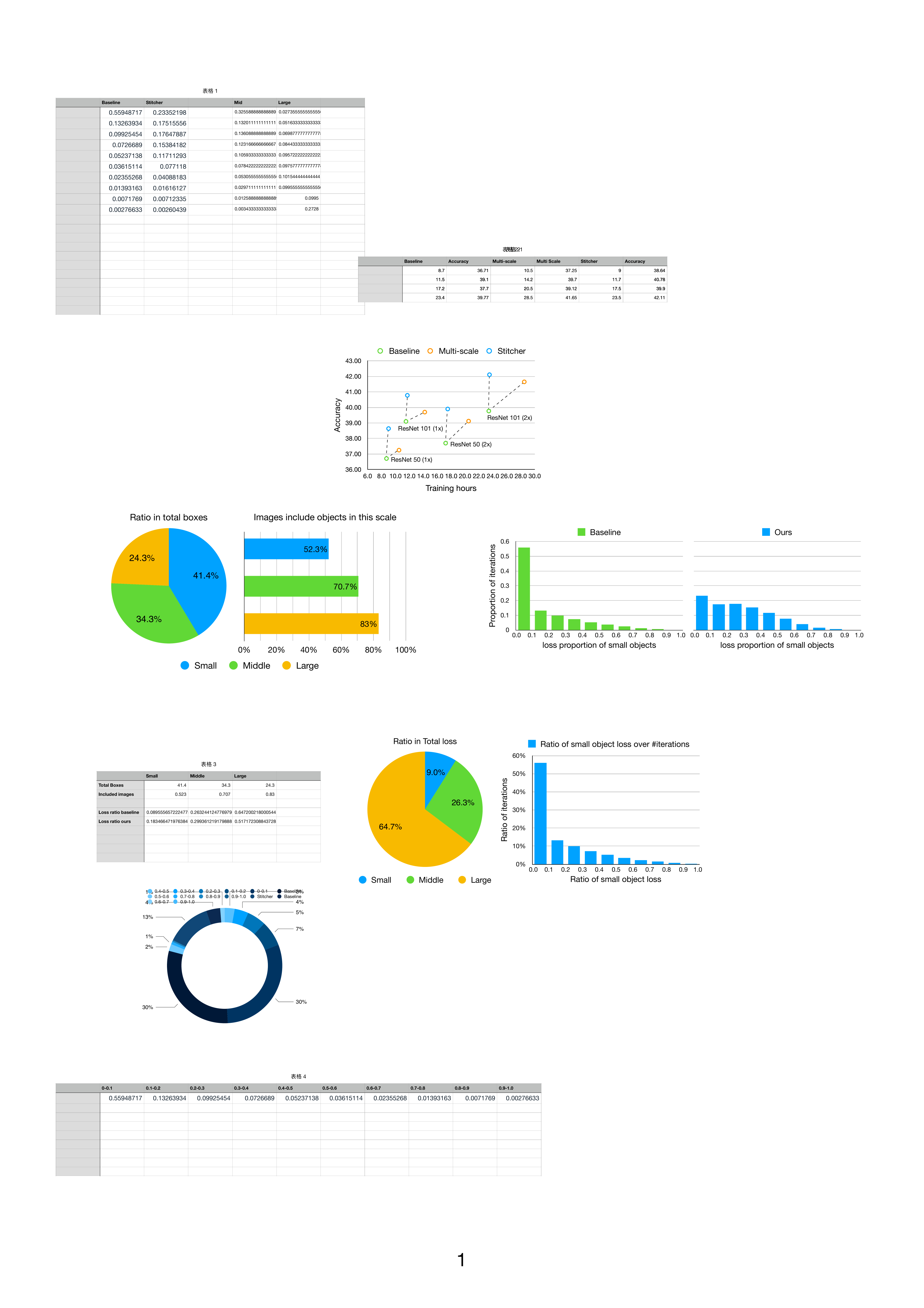}
    \caption{Loss proportion of minority scales before and after.}
    \label{fig:loss_distribution2}
\end{figure}

Beyond the overall observation, we also investigate into loss proportions of the small scales. As shown in the Figure~\ref{fig:loss_distribution2} left, more than half of the training iterations undergo an extremely low loss proportion of small objects, dubbed 0.1. By adopting our proposed method, the scale variation from the perspective of loss proportion distribution get much alleviated (see Figure~\ref{fig:loss_distribution2} right).

\subsection{Other Merits}
Beyond the performance enhancement, there are also extra merits brought by the proposed dynamic scale training during scale variation handling.
\subsubsection{Speed-Accuracy Trade-off}
We find an improvement upon speed-accuracy trade-off as shown in Table~\ref{tab:speed-accuracy-trade-off}. It could be observed that our method runs on par with the baseline (AP: 37.0 \textit{vs.} 36.7) given inputs of much smaller sizes (resolution: (512, 853) \textit{vs.} (800, 1333)) and meanwhile is 1.6$\times$ faster.


\begin{table}[htbp]
\caption{Speed-accuracy trade-off merit brought by dynamic scale training. The baseline is Faster R-CNN with ResNet-50 and FPN.}

\label{tab:speed-accuracy-trade-off}
\centering
\resizebox{0.8\linewidth}{!}{
\begin{tabular}{l|c|c|l}
\toprule
        &  Resolution    &   Inference time  & AP \\ \midrule
Baseline & (800, 1333) & 56 ms / img & 36.7 \\
Baseline & (512, 853) & 35 ms / img & 33.5 \\
Ours & (800, 1333) & 56 ms / img & 38.6 \\ 
Ours & (512, 853) & 35 ms / img & 37.0 \\ \bottomrule
\end{tabular}}
\end{table}

\subsubsection{Fast Convergence}
We discover the fast convergence capacity of our proposed DST method. Referring to Table~\ref{tab:intro_table}, after applying DST, it nearly halves (iters: 50k vs. 90k) the training iterations to achieve the same accuracy to the baseline.

\begin{table}[t]
    \caption{Fast convergence merit brought dynamic scale training. The baseline is Faster R-CNN with ResNet-50 and FPN.}
    
    \label{tab:intro_table}
    \centering
    \resizebox{0.8\linewidth}{!}{
    \begin{tabular}{l|c|l|lll}
    \toprule
                       & Iterations & AP & AP$_s$ & AP$_m$ & AP$_l$ \\ \midrule
    Baseline & 90k     & 36.7                    & 21.1                        & 39.9                        & 48.1                       \\
    Ours               & \textbf{50k}       & 36.7                    & 22.9                       & 39.9                        & 46.6                       \\
    Ours               & 90k        & 38.6             & 24.4                       & 41.9                        & 49.3                       \\ \bottomrule
    \end{tabular}
    }
\end{table}

\subsection{Corner cases of collage}
\label{subsec:corner_cases}
Recalling the collage procedure, regular images are down-scaled before being stitched to form the collage components. This might produce extremely tiny objects more likely to be a noisy pattern (from existing small objects). To investigate the impact, we discard tiny samples whose box areas less than 100 pixels. Before removal, the results are AP: 38.6, AP$_s$: 24.4, AP$_m$: 41.9, AP$_l$: 49.3. After removal, we obtain AP: 38.6, AP$_s$: 24.7, AP$_m$: 41.8, AP$_l$: 49.1. This demonstrates that tiny patterns do not affect the overall performance but might hamper the quality on small scales.


\section{Conclusion}
In this paper, we propose a simple yet effective \textit{dynamic scale training method (DST)} for object detection. By relieving the scale variation issue in virtue of feedback information from the optimization process, we observe significant gains in detection performance. Moreover, it introduces efficient convergence during training and does not affect the inference time as a free lunch. Abundant experiments have been conducted to verify its efficacy on various backbones, training periods, datasets, and different tasks. DST could be easily incorporated into modern detectors and steadily enhances the detection quality. We expect it could serve as a common configuration in the future, facilitating further dynamic training research for object detection.

\newpage{}
{\small
\bibliographystyle{ieee_fullname}
\bibliography{egbib}

\begin{thebibliography}{10}\itemsep=-1pt

\bibitem{imagepyramid}
Edward~H Adelson, Charles~H Anderson, James~R Bergen, Peter~J Burt, and Joan~M
  Ogden.
\newblock Pyramid methods in image processing.
\newblock {\em RCA engineer}, 29(6):33--41, 1984.

\bibitem{yolov4}
Alexey Bochkovskiy, Chien{-}Yao Wang, and Hong{-}Yuan~Mark Liao.
\newblock Yolov4: Optimal speed and accuracy of object detection.
\newblock {\em CoRR}, abs/2004.10934, 2020.

\bibitem{soft-nms}
Navaneeth Bodla, Bharat Singh, Rama Chellappa, and Larry~S. Davis.
\newblock Soft-nms - improving object detection with one line of code.
\newblock In {\em ICCV}, pages 5562--5570, 2017.

\bibitem{lapnet}
Florian Chabot, Mohamed Chaouch, and Quoc~Cuong Pham.
\newblock Lapnet: Automatic balanced loss and optimal assignment for real-time
  dense object detection.
\newblock {\em CoRR}, abs/1911.01149, 2019.

\bibitem{autoaug}
Ekin~Dogus Cubuk, Barret Zoph, Dandelion Man{\'{e}}, Vijay Vasudevan, and
  Quoc~V. Le.
\newblock Autoaugment: Learning augmentation policies from data.
\newblock In {\em CVPR}, pages 113--123, 2019.

\bibitem{deformable}
Jifeng Dai, Haozhi Qi, Yuwen Xiong, Yi Li, Guodong Zhang, Han Hu, and Yichen
  Wei.
\newblock Deformable convolutional networks.
\newblock In {\em ICCV}, pages 764--773, 2017.

\bibitem{PascalVOC}
Mark Everingham, S.~M.~Ali Eslami, Luc J.~Van Gool, Christopher K.~I. Williams,
  John~M. Winn, and Andrew Zisserman.
\newblock The pascal visual object classes challenge: {A} retrospective.
\newblock {\em International Journal of Computer Vision}, 111(1):98--136, 2015.

\bibitem{nas-fpn}
Golnaz Ghiasi, Tsung-Yi Lin, and Quoc~V Le.
\newblock Nas-fpn: Learning scalable feature pyramid architecture for object
  detection.
\newblock In {\em CVPR}, pages 7036--7045, 2019.

\bibitem{fast-r-cnn}
Ross~B. Girshick.
\newblock Fast {R-CNN}.
\newblock In {\em ICCV}, pages 1440--1448, 2015.

\bibitem{R-CNN}
Ross~B. Girshick, Jeff Donahue, Trevor Darrell, and Jitendra Malik.
\newblock Rich feature hierarchies for accurate object detection and semantic
  segmentation.
\newblock In {\em CVPR}, pages 580--587, 2014.

\bibitem{he2019rethinking}
Kaiming He, Ross Girshick, and Piotr Doll{\'a}r.
\newblock Rethinking imagenet pre-training.
\newblock In {\em ICCV}, pages 4918--4927, 2019.

\bibitem{maskrcnn}
Kaiming He, Georgia Gkioxari, Piotr Doll{\'a}r, and Ross Girshick.
\newblock Mask r-cnn.
\newblock In {\em ICCV}, pages 2961--2969, 2017.

\bibitem{he2016deep}
Kaiming He, Xiangyu Zhang, Shaoqing Ren, and Jian Sun.
\newblock Deep residual learning for image recognition.
\newblock In {\em CVPR}, pages 770--778, 2016.

\bibitem{ioffe2015batch}
Sergey Ioffe and Christian Szegedy.
\newblock Batch normalization: Accelerating deep network training by reducing
  internal covariate shift.
\newblock {\em CoRR}, abs/1502.03167, 2015.

\bibitem{MAL}
Wei Ke, Tianliang Zhang, Zeyi Huang, Qixiang Ye, Jianzhuang Liu, and Dong
  Huang.
\newblock Multiple anchor learning for visual object detection.
\newblock In {\em CVPR}, pages 10206--10215, 2020.

\bibitem{tridentnet}
Yanghao Li, Yuntao Chen, Naiyan Wang, and Zhaoxiang Zhang.
\newblock Scale-aware trident networks for object detection.
\newblock 2019.

\bibitem{fpn}
Tsung{-}Yi Lin, Piotr Doll{\'{a}}r, Ross~B. Girshick, Kaiming He, Bharath
  Hariharan, and Serge~J. Belongie.
\newblock Feature pyramid networks for object detection.
\newblock In {\em CVPR}, pages 936--944, 2017.

\bibitem{retinanet}
Tsung{-}Yi Lin, Priya Goyal, Ross~B. Girshick, Kaiming He, and Piotr
  Doll{\'{a}}r.
\newblock Focal loss for dense object detection.
\newblock In {\em ICCV}, pages 2999--3007, 2017.

\bibitem{coco}
Tsung-Yi Lin, Michael Maire, Serge Belongie, James Hays, Pietro Perona, Deva
  Ramanan, Piotr Doll{\'a}r, and C~Lawrence Zitnick.
\newblock Microsoft coco: Common objects in context.
\newblock In {\em ECCV}, pages 740--755, 2014.

\bibitem{ASFF}
Songtao Liu, Di Huang, and Yunhong Wang.
\newblock Learning spatial fusion for single-shot object detection.
\newblock {\em CoRR}, abs/1911.09516, 2019.

\bibitem{panet}
Shu Liu, Lu Qi, Haifang Qin, Jianping Shi, and Jiaya Jia.
\newblock Path aggregation network for instance segmentation.
\newblock In {\em CVPR}, pages 8759--8768, 2018.

\bibitem{ssd}
Wei Liu, Dragomir Anguelov, Dumitru Erhan, Christian Szegedy, Scott~E. Reed,
  Cheng{-}Yang Fu, and Alexander~C. Berg.
\newblock {SSD:} single shot multibox detector.
\newblock In {\em ECCV}, pages 21--37, 2016.

\bibitem{hambox}
Yang Liu, Xu Tang, Xiang Wu, Junyu Han, Jingtuo Liu, and Errui Ding.
\newblock Hambox: Delving into online high-quality anchors mining for detecting
  outer faces.
\newblock {\em CoRR}, abs/1912.09231, 2019.

\bibitem{ming2019group}
Xiang Ming, Fangyun Wei, Ting Zhang, Dong Chen, and Fang Wen.
\newblock Group sampling for scale invariant face detection.
\newblock In {\em CVPR}, pages 3446--3456, 2019.

\bibitem{megdet}
Chao Peng, Tete Xiao, Zeming Li, Yuning Jiang, Xiangyu Zhang, Kai Jia, Gang Yu,
  and Jian Sun.
\newblock Megdet: A large mini-batch object detector.
\newblock In {\em CVPR}, pages 6181--6189, 2018.

\bibitem{peng2019pod}
Junran Peng, Ming Sun, Zhaoxiang Zhang, Tieniu Tan, and Junjie Yan.
\newblock Pod: practical object detection with scale-sensitive network.
\newblock In {\em CVPR}, pages 9607--9616, 2019.

\bibitem{faster-rcnn}
Shaoqing Ren, Kaiming He, Ross~B. Girshick, and Jian Sun.
\newblock Faster {R-CNN:} towards real-time object detection with region
  proposal networks.
\newblock {\em {IEEE} Trans. Pattern Anal. Mach. Intell.}, 39(6):1137--1149,
  2017.

\bibitem{ohem}
Abhinav Shrivastava, Abhinav Gupta, and Ross~B. Girshick.
\newblock Training region-based object detectors with online hard example
  mining.
\newblock In {\em CVPR}, pages 761--769, 2016.

\bibitem{SNIP}
Bharat Singh and Larry~S. Davis.
\newblock An analysis of scale invariance in object detection {\-} {SNIP}.
\newblock In {\em CVPR}, pages 3578--3587, 2018.

\bibitem{SNIPER}
Bharat Singh, Mahyar Najibi, and Larry~S. Davis.
\newblock {SNIPER:} efficient multi-scale training.
\newblock In {\em NeurIPS}, pages 9333--9343, 2018.

\bibitem{tian2019fcos}
Zhi Tian, Chunhua Shen, Hao Chen, and Tong He.
\newblock Fcos: Fully convolutional one-stage object detection.
\newblock In {\em CVPR}, pages 9627--9636, 2019.

\bibitem{wu2018group}
Yuxin Wu and Kaiming He.
\newblock Group normalization.
\newblock In {\em ECCV}, pages 3--19, 2018.

\bibitem{resnext}
Saining Xie, Ross~B. Girshick, Piotr Doll{\'{a}}r, Zhuowen Tu, and Kaiming He.
\newblock Aggregated residual transformations for deep neural networks.
\newblock In {\em CVPR}, pages 5987--5995, 2017.

\bibitem{dynamic-rcnn}
Hongkai Zhang, Hong Change, Bingpeng Ma, Naiyan Wang, and Xilin Chen.
\newblock Dynamic r-cnn: Towards high quality object detection via dynamic
  training.
\newblock In {\em ECCV}, 2020.

\bibitem{atss}
Shifeng Zhang, Cheng Chi, Yongqiang Yao, Zhen Lei, and Stan~Z. Li.
\newblock Bridging the gap between anchor-based and anchor-free detection via
  adaptive training sample selection.
\newblock In {\em CVPR}, pages 9759--9768, 2020.

\bibitem{freeanchor}
Xiaosong Zhang, Fang Wan, Chang Liu, Rongrong Ji, and Qixiang Ye.
\newblock Freeanchor: Learning to match anchors for visual object detection.
\newblock In {\em NIPS}, pages 147--155, 2019.

\bibitem{cheap-pretrain}
Dongzhan Zhou, Xinchi Zhou, Hongwen Zhang, Shuai Yi, and Wanli Ouyang.
\newblock Cheaper pre-training lunch: An efficient paradigm for object
  detection.
\newblock In {\em ECCV}, 2020.

\bibitem{FSAF}
Chenchen Zhu, Yihui He, and Marios Savvides.
\newblock Feature selective anchor-free module for single-shot object
  detection.
\newblock In {\em CVPR}, pages 840--849, 2019.

\bibitem{autoaugfordet}
Barret Zoph, Ekin~D. Cubuk, Golnaz Ghiasi, Tsung{-}Yi Lin, Jonathon Shlens, and
  Quoc~V. Le.
\newblock Learning data augmentation strategies for object detection.
\newblock In {\em CVPR}, page 770–778, 2019.

\end{thebibliography}
}
\end{document}